# The Hierarchical Dirichlet Process Hidden Semi-Markov Model

**Matthew J. Johnson**
Department of EECS
Massachusetts Institute of Technology
Cambridge, MA 02139
mattjj@csail.mit.edu

**Alan S. Willsky**
Department of EECS
Massachusetts Institute of Technology
Cambridge, MA 02139
willsky@mit.edu

## Abstract

There is much interest in the Hierarchical Dirichlet Process Hidden Markov Model (HDP-HMM) as a natural Bayesian nonparametric extension of the traditional HMM. However, in many settings the HDP-HMM's strict Markovian constraints are undesirable, particularly if we wish to learn or encode non-geometric state durations. We can extend the HDP-HMM to capture such structure by drawing upon explicit-duration semi-Markovianity, which has been developed in the parametric setting to allow construction of highly interpretable models that admit natural prior information on state durations.

In this paper we introduce the explicit-duration HDP-HSMM and develop posterior sampling algorithms for efficient inference in both the direct-assignment and weak-limit approximation settings. We demonstrate the utility of the model and our inference methods on synthetic data as well as experiments on a speaker diarization problem and an example of learning the patterns in Morse code.

## 1 Introduction

Given a set of sequential data in an unsupervised setting, we often aim to infer meaningful states, or "topics," present in the data along with characteristics that describe and distinguish those states. For example, in a speaker diarization (or who-spoke-when) problem, we are given a single audio recording of a meeting and wish to infer the number of speakers present, when they speak, and some characteristics governing their speech patterns. In analyzing DNA sequences, we may want to identify and segment region types using prior knowledge about region length distributions.

Such learning problems for sequential data are pervasive, and so we would like to build general models that are both flexible enough to be applicable to many domains and expressive enough to encode the appropriate information.

Hidden Markov Models (HMMs) have proven to be excellent general models for approaching such learning problems in sequential data, but they have two significant disadvantages: (1) state duration distributions are necessarily restricted to a geometric form that is not appropriate for many real-world data, and (2) the number of hidden states must be set a priori so that model complexity is not inferred from data in a Bayesian way.

Recent work in Bayesian nonparametrics has addressed the latter issue. In particular, the Hierarchical Dirichlet Process HMM (HDP-HMM) has provided a powerful framework for inferring arbitrarily large state complexity from data [9]. However, the HDP-HMM does not address the issue of non-Markovianity in real data. The Markovian disadvantage is even compounded in the nonparametric setting, since non-Markovian behavior in data can lead to the creation of unnecessary extra states and unrealistically rapid switching dynamics [3].

One approach to avoiding the rapid-switching problem is the Sticky HDP-HMM [2], which introduces a learned self-transition bias to discourage rapid switching. The Sticky model has demonstrated significant performance improvements over the HDP-HMM for several applications [3]. However, it shares the HDP-HMM's restriction to geometric state durations, thus limiting the model's expressiveness regarding duration structure. Moreover, its global self-transition bias is shared amongst all states, and so it does not allow for learning state-specific duration information. The infinite Hierarchical HMM [5] induces non-Markovian state durations at the coarser levels of its state hierarchy, but even the coarser levels are constrained to have a sum-of-geometrics form, and hence it can be difficult to incorporate prior information.

These potential improvements to the HDP-HMM motivate an investigation of explicit-duration semi-Markovianity, which has a history of success in the parametric setting. In this paper, we combine semi-Markovian ideas with the HDP-HMM to construct a general class of models that allow for both Bayesian nonparametric inference of state complexity as well as incorporation of general duration distributions. In addition, the sampling techniques we develop for the Hierarchical Dirichlet Process Hidden semi-Markov Model (HDP-HSMM) provide new approaches to inference in HDP-HMMs that can avoid some of the difficulties which result in slow mixing rates.

In Section 2, we provide a brief introduction to both the HDP-HMM and parametric Hidden semi-Markov models. In Section 3 we introduce the HDP-HSMM, and in Section 4 we develop efficient sampling inference methods including a direct-assignment sampler and a weak-limit approximate sampler. Section 5 demonstrates the effectiveness of the HDP-HSMM model on both synthetic and real data, including applications to a speaker diarization problem and to learning the structure of Morse code.

Code and supplementary materials can be found at www.mit.edu/~mattjj/uai2010.

## 2 Background

### 2.1 HDP-HMM Background

The HDP-HMM [9] provides a natural Bayesian nonparametric treatment of the classical Hidden Markov Model approach to sequential statistical modeling. The model employs an HDP prior over an infinite state space, which enables both inference of state complexity and Bayesian mixing over models of varying complexity. Thus the HDP-HMM subsumes the usual model selection problem, replacing other techniques for choosing a fixed number of HMM states such as cross-validation procedures, which can be computationally expensive and restrictive. Furthermore, the HDP-HMM inherits many of the desirable properties of the HDP prior, especially the ability to encourage model parsimony while allowing complexity to grow with the number of observations. We provide a brief overview of the HDP-HMM model and relevant inference techniques, which we extend to develop the HDP-HSMM.

The generative HDP-HMM model (Figure 1) can be

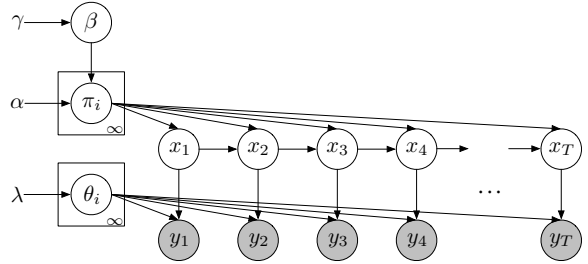

Figure 1: Graphical model for the HDP-HMM.

summarized as:

$$\beta|\gamma \sim \text{GEM}(\gamma)$$
$$\pi_j|\beta, \alpha \sim \text{DP}(\alpha, \beta) \qquad j = 1, 2, \ldots$$
$$\theta_j|H, \lambda \sim H(\lambda) \qquad j = 1, 2, \ldots$$
$$x_t|\{\pi_j\}_{j=1}^\infty, x_{t-1} \sim \pi_{x_{t-1}} \qquad t = 1, \ldots, T$$
$$y_t|\{\theta_j\}_{j=1}^\infty, x_t \sim f(\theta_{x_t}) \qquad t = 1, \ldots, T$$

where GEM denotes a stick-breaking process [8].

The variable sequence $(x_t)$ represents the hidden state sequence, and $(y_t)$ represents the observation sequence drawn from the observation distribution $f$. The set of state-specific observation distribution parameters is represented by $\{\theta_j\}$, which are draws from the prior $H$ parameterized by $\lambda$. The HDP plays the role of a prior over infinite transition matrices: each $\pi_j$ is a DP draw and is interpreted as the transition distribution from state $j$, i.e. the $j$th row of the transition matrix. The $\pi_j$ are linked by being DP draws parameterized by the same discrete measure $\beta$, thus $E[\pi_j] = \beta$ and the transition distributions tend to have their mass concentrated around a typical set of states, providing the desired bias towards re-entering and re-using a consistent set of states.

Though several of the objects in the HDP-HMM are infinite and thus cannot be fully instantiated, the construction allows us to analytically marginalize over $\beta$, $\{\pi_j\}$, and $\{\theta_j\}$ (given $H$ is conjugate to $f$) and deal only with the $(x_t)$ variables. Using this technique we can effectively represent the HDP by only instantiating as much as we need: the $x_t$ become Markov exchangeable, and we can exploit the exchangeability to produce samplers of the form of the Chinese Restaurant Franchise (CRF) [9].

The CRF sampling methods provide us with effective approximate inference for the full infinite-dimensional HDP, but they have a particular weakness in the context of the HDP-HMM: each state transition must be re-sampled individually, and strong correlations within the state sequence significantly reduce mixing rates for such operations [3].

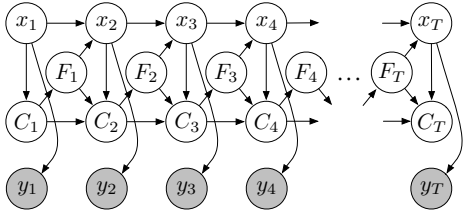

Figure 2: A graphical model for the HSMM with explicit counter and finish nodes.

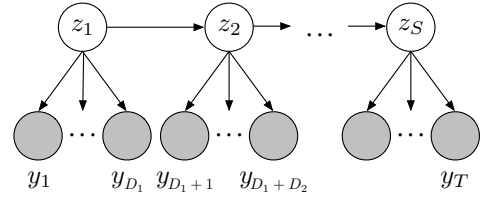

Figure 3: HSMM interpreted as a Markov chain on a set of super-states, $(z_s)_{s=1}^{S}$. The number of shaded nodes associated with each $z_s$ is random, drawn from a state-specific duration distribution.

As a result, finite approximations to the HDP have been studied for the purpose of providing alternative approximate inference schemes. Of particular note is the popular weak limit approximation, used in [2], which has been shown to reduce mixing times for HDP-HMM inference while sacrificing little of the "tail" of the infinite transition matrix.

## 2.2 HSMM Background

There are several modeling approaches to semi-Markovianity [6], but here we focus on *explicit duration* semi-Markovianity; i.e., we are interested in the setting where each state's duration is given an explicit distribution.

The basic idea underlying this HSMM formalism is to augment the generative process of a standard HMM with a random state duration time, drawn from some state-specific distribution when the state is entered. The state remains constant until the duration expires, at which point there is a Markov transition to a new state. It can be cumbersome to draw the process into a proper graphical model, but one compelling representation in [6] is to add explicit "timer" and "finish" variables, as depicted in Figure 2. The $(C_t)$ variables serve to count the remaining duration times, and are deterministically decremented to zero. The $(F_t)$ variables indicate that there is a Markov transition at time $t+1$, and $F_t = 1$ causes $C_{t+1}$ to be sampled from the duration distribution of $x_{t+1}$.

An equivalent and somewhat more intuitive picture is given in Figure 3 (also from [6]), though the number of nodes in the model is itself random. In this picture, we see there is a standard Markov chain on "super-state" nodes, $(z_s)_{s=1}^{S}$, and these super-states in turn emit random-length segments of observations, of which we observe the first $T$. The symbol $D_i$ is used to denote the random length of the observation segment of super-state $i$ for $i = 1, \ldots, S$. The "super-state" picture separates the Markovian transitions from the segment durations, and is helpful in building an effective CRF-style sampler for the HDP-HSMM.

It is often taken as convention that state self-transitions should be ruled out in an HSMM, because if a state can self-transition then the duration distribution does not fully capture a state's possible duration length. We adopt this convention, which has a significant impact on the inference algorithms described in Section 4. When defining an HSMM model, one must also choose whether the observation sequence ends exactly on a segment boundary or whether the observations are *censored* at the end, so that the final segment may possibly be cut off in the observations. This censoring convention allows for slightly simpler formulae and computations, and thus is adopted in this paper. We do, however, assume the observations begin on a segment boundary. For more details and alternative conventions, see [4].

It is possible to perform efficient message-passing inference along an HSMM state chain (conditioned on parameters and observations) in a way similar to the standard alpha-beta dynamic programming algorithm for standard HMMs. The "backwards" messages are crucial in the development of efficient sampling inference in Section 4, and so we briefly describe the relevant part of the existing HSMM message-passing algorithm. As derived in [6], we can define and compute the backwards message from $t$ to $t+1$ as:

$$\beta_t(i) \triangleq p(y_{t+1:T}|x_t = i, F_t = 1) \qquad (1)$$
$$= \sum_j \beta_t^*(j) p(x_{t+1} = j | x_t = i)$$
$$\beta_t^*(i) \triangleq p(y_{t+1:T}|x_{t+1} = i, F_t = 1)$$
$$= \sum_{d=1}^{T-t} \beta_{t+d}(i) \overbrace{p(D_{t+1} = d | x_{t+1} = i)}^{\text{duration prior term}}$$
$$\cdot \underbrace{p(y_{t+1:t+d}|x_{t+1} = i, D = d)}_{\text{likelihood term}}$$
$$\beta_T(i) \triangleq 1$$

where we have split the messages into $\beta$ and $\beta^*$ components for convenience and used $y_{k_1:k_2}$ to denote $(y_{k_1}, \ldots, y_{k_2})$. Also note that we have used $D_{t+1}$ to represent the duration of the segment beginning at time $t+1$. The conditioning on the parameters of the distributions is suppressed from the notation.

The $F_t = 1$ condition indicates a new segment begins at $t+1$, and so to compute the message from $t+1$ to $t$ we sum over all possible lengths $d$ for the segment beginning at $t+1$, using the backwards message at $t+d$ to provide aggregate future information given a boundary just after $t+d$. There is an additional censoring term in the expression for $\beta_t^*(i)$ which is not shown for simplicity; it is described in [4].

The greater expressivity of the HSMM model necessarily increases the computational cost of the message passing algorithm: the above message passing requires $\mathcal{O}(T^2N + TN^2)$ basic operations for a chain of length $T$ and state cardinality $N$, while the corresponding HMM message passing algorithm requires only $\mathcal{O}(TN^2)$. However, if we truncate possible segment lengths included in the inference messages to some maximum $d_{\max}$, we can instead express the asymptotic message passing cost as $\mathcal{O}(Td_{\max}N^2)$. Such truncations are often natural because both the duration prior term and the segment likelihood term contribute to the product rapidly vanishing with sufficiently large $d$. Though the increased complexity of message-passing over an HMM significantly increases the cost per iteration of sampling inference, the cost is offset because HSMM samplers often require far fewer total iterations to converge (see Section 4).

## 3 Defining the HDP-HSMM

In this section we formally define the HDP-HSMM and point out some particular details in the definition that have significant implications for inference algorithms. The generative process of the HDP-HSMM employs a combination of the preceding ideas:

$$\beta | \gamma \sim \text{GEM}(\gamma)$$
$$\pi_j | \beta, \alpha \sim \text{DP}(\alpha, \beta) \qquad j = 1, 2, \ldots$$
$$\theta_j | H, \lambda \sim H(\lambda) \qquad j = 1, 2, \ldots$$
$$\omega_j | \Omega \sim \Omega \qquad j = 1, 2, \ldots$$

$\tau := 0$, $s := 1$, while $\tau < T$ do:
$$z_s | \{\pi_j\}_{j=1}^\infty, z_{s-1} \sim \tilde{\pi}_{z_{s-1}}$$
$$D_s | \omega \sim D(\omega_{z_s})$$
$$y_s = y_{\tau+1:\tau+D_s+1} | \{\theta_j\}_{j=1}^\infty, z_s, D_s \overset{\text{iid}}{\sim} f(\theta_{z_s})$$
$$\tau := \tau + D_s$$
$$s := s + 1$$

where we have used $(z_s)$ as a super-state sequence indexed by $s$ and $\{\omega_j\}_{j=1}^\infty$ to represent the parameters for the duration distributions of each of the states, with $D$ representing the class of duration distributions. At the end of the process, we censor the observations to have length $T$ exactly, cutting off any excess observations if necessary, so as to generate $y_{1:T}$. It is also convenient to refer to $x_t$ as the state index to which an observation $y_t$ belongs, for $t = 1, \ldots, T$.

Note that we draw $z_s | \{\pi_j\}, z_{s-1}$ from $\tilde{\pi}_{z_{s-1}}$, which we use to denote the conditional measure constructed from $\pi_{z_{s-1}}$ by removing the atom corresponding to $z_{s-1}$ and re-normalizing appropriately. This part of the construction effectively rules out self-transitions.

If $D$, the duration distribution class, is geometric, we effectively recover the HDP-HMM (just as we would recover a standard HMM from an HSMM with geometric duration distributions) but the resulting inference procedure remains distinct from the HDP-HMM. Thus the HDP-HSMM sampling inference methods described in the next section provide a novel alternative to existing HDP-HMM samplers with some potentially significant advantages.

## 4 Sampling Inference

In this section we briefly introduce both a new direct-assignment sampler (of the style in [9]) and a new weak-limit sampler (similar to that in [3]) for efficient HDP-HSMM inference. For details, see the supplementary material.

### 4.1 Finite HSMM Blocked State Sampling

It is important for the development of both samplers to describe a message-backwards, sample-forwards technique to block sample from the posterior distribution on a (finite-dimensional) semi-Markov state chain. A similar idea [3] is employed to block-sample the entire hidden state sequence in finite (or weak-limit HDP) HMM models. However, in the case of HSMMs we must also sample from the posterior of the duration distribution.

If we compute the backwards messages $\beta$ and $\beta^*$ of (1), then we can easily draw a posterior sample for the first state according to:

$$p(x_1 = i | y_{1:T}) \propto p(x_1 = i) p(y_{1:T} | x_1 = i, F_0 = 1)$$
$$= p(x_1 = i) \beta_0^*(i)$$

where we have used the assumption that the observation sequence begins on a segment boundary ($F_0 = 1$), and again we have suppressed notation for conditioning on parameters. Conditioning on the initial state draw, $\bar{x}_1$, we can then draw a sample of $D_1 | y_{1:T}, x_1,$

the posterior duration of the first state, via:

$$p(D_1 = d|y_{1:T}, x_1 = \bar{x}_1, F_0 = 1) = \qquad (2)$$
$$\frac{p(D_1 = d)p(y_{1:d}|D_1 = d, x_1 = \bar{x}_1, F_0 = 1)\beta_d(\bar{x}_1)}{\beta_0^*(\bar{x}_1)}$$

We can repeat the process by then considering $x_{D_1+1}$ to be our new initial state with initial distribution given by $p(x_{D_1+1} = i|x_1 = \bar{x}_1)$. This forward-sampling algorithm with posterior draws from duration distributions has not previously been described for the HSMM. It can be viewed as an extension of the changepoint sampling technique developed in [1] (in which segment parameters are assumed independent) to the setting where segment parameters are revisited according to Markovian state dynamics.

### 4.2 Direct-Assignment Sampler

To create a direct-assignment sampler based on the HDP-HMM direct-assignment sampler of [9], we can leverage the viewpoint of an HSMM as an HMM on super-state segments and split the sampling update into two steps. First, conditioning on a segmentation (which defines super-state boundaries but not labels), we can view blocks of observations as atomic with a single predictive likelihood score for the entire block. We can then run an HDP-HMM direct-assignment sampler on the super-state chain with the caveat that we have outlawed self-transitions. Second, given a super-state sequence we can efficiently re-sample the segmentation boundaries.

We can rule out self-transitions in the super-state sequence while maintaining a "complete" set of transitions by performing latent history sampling [7]. We sample super-state transitions without any constraints, and we reject any samples that result in self-transitions while counting the number of such rejections for each state. These "dummy" self-transitions, which are not reflected in the super-state sequence, allow us to sample posterior super-state transitions according to the standard HDP direct-assignment sampler.

To re-sample the segmentation given a super-state sequence, we first sample the posterior observation parameters (which are marginalized in first sampling step) and duration parameters (which are independent of the first step) for each unique state. Next, we index the super-state sequence in order from 1 to $S$ and construct an $S$-state finite HSMM on states $(x_t)_1^T$ with a transition matrix that has 1s on its first superdiagonal and zeros elsewhere. We define the observation distribution of state $s \in \{1, \ldots, S\}$ to be $f(\theta_{z_s})$ and the duration distribution of state $s$ to be $D(\omega_{z_s})$. If we set $x_1 = z_1$, this constructed HSMM always follows the super-state sequence upon which we are conditioning, and so we can run the finite HSMM's messages-backwards, sample-forwards scheme to efficiently construct a posterior sample of the segmentation.

It is interesting to consider how this sampling algorithm differs from the HDP-HMM algorithm when geometric duration distributions are used. From a generative standpoint the model classes are identical, but in the HDP-HSMM sampling algorithm entire state segments are resampled at once. Thus the HDP-HSMM sampling method may be useful not only for the case of non-geometric durations, but also as an HDP-HMM sampler that avoids the usual mixing issues.

### 4.3 Weak-Limit Sampler

The weak-limit sampler for an HDP-HMM [2] constructs a finite approximation to the HDP transitions prior with finite $L$-dimensional Dirichlet distributions, motivated by the fact that the infinite limit of such a construction converges in distribution to a true HDP:

$$\beta|\gamma \sim \text{Dir}(\gamma/L, \ldots, \gamma/L)$$
$$\pi_j|\alpha, \beta \sim \text{Dir}(\alpha\beta_1, \ldots, \alpha\beta_L) \qquad j = 1, \ldots, L$$

where we again interpret $\pi_j$ as the transition distribution for state $j$ and $\beta$ as the distribution which ties state distributions together and encourages shared sparsity. Practically, the weak limit approximation enables the instantiation of the transition matrix in a finite form, and thus allows block sampling of the hidden state chain, resulting in greatly accelerated mixing.

The main challenge is that, as in the infinite case, the hierarchical Dirichlet construction is not a conjugate prior for the transitions present in the hidden state sequence because we do not observe any self-transitions. However, we can sample geometric auxiliary variables to complete the data, effectively marginalizing over self-transitions and allowing conjugate inference. See the supplementary material for details.

## 5 Results

In this section we apply the HDP-HSMM weak-limit sampler to both synthetic and real data.

### 5.1 Synthetic Data

We evaluated the HDP-HSMM model and inference techniques by generating observations from both HSMMs and HMMs and comparing performance to the HDP-HMM. The models learn many parameters including observation, duration, and transition parameters for each state. For the sake of brevity we present the normalized Hamming error (as described in [2]) of the sampled state sequences as a summary metric, since it involves all learned parameters. To compute the normalized Hamming error, we greedily identify inferred state labels with the ground truth labels and measure the proportion of correctly labeled states. Note that if any states are superfluous or missing in the inferred sequence their corresponding labels are counted purely as error. In these plots, the blue line indicates the median error across all 25 chains, while the red lines indicate 10th and 90th percentile errors.

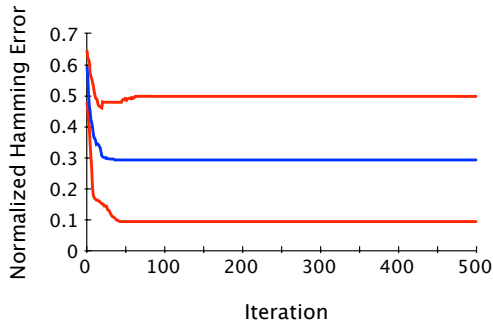

(a) HDP-HMM

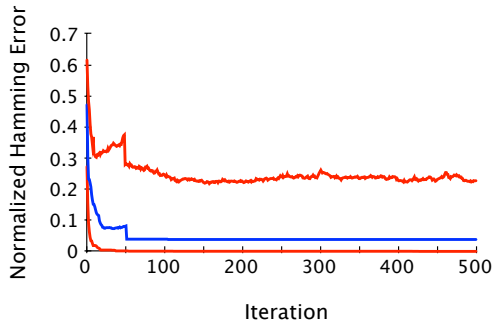

(b) HDP-HSMM

Figure 4: HDP-HSMM and HDP-HMM applied to data from a Poisson-HSMM.

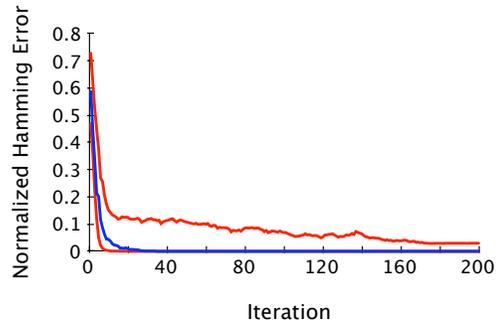

(a) HDP-HMM

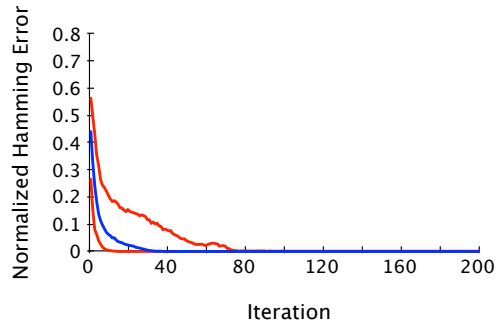

(b) HDP-HSMM

Figure 5: The HDP-HSMM and HDP-HMM applied to data from an HMM.

Figure 4 summarizes the results of applying both an HDP-HSMM and an HDP-HMM to data generated from an HSMM with four states and Poisson durations. The observations for each state are mixtures of 2-dimensional Gaussians with significant overlap, with parameters for each state sampled i.i.d. from a Normal Inverse-Wishart (NIW) prior. There were 25 chains run in the experiment, with 5 chains on each of 5 generated observation sequences. The HDP-HMM is unable to capture the proper duration statistics and so its state sampling error remains high, while the HDP-HSMM is able to effectively capture the correct temporal model and thus effectively separate the states and significantly reduce posterior uncertainty. The HDP-HMM also fails to identify the true number of states.

By setting the class of duration distributions to be a strict superclass of the geometric distribution, we can allow an HDP-HSMM model to learn an HMM from data when appropriate. One such distribution class is the class of negative binomial distributions, denoted $\text{NegBin}(r,p)$, the discrete analog of the Gamma distribution, which covers the class of geometric distributions when $r = 1$.

Figure 5 shows a negative binomial HDP-HSMM learning an HMM model from data generated from an HMM with four states. The observation distribution for each state is a 10-dimensional Gaussian, again with parameters sampled i.i.d. from a NIW prior. The prior over $r$ was set to be uniform on $\{1, 2, \ldots, 6\}$. The sampler chains quickly concentrated at $r = 1$ for all state duration distributions. There is only a slight loss in mixing time for the negative binomial HDP-HSMM compared to the HDP-HMM on this data.

### 5.2 Learning Morse Code

As an example of duration information disambiguating states, we also applied both an HDP-HSMM and an HDP-HMM to spectrogram data from audio of the Morse code alphabet (see Figure 6). The data can clearly be partitioned into "tone" and "silence" clus-ters without inspecting temporal structure, but only by incorporating duration information can we disambiguate the "short tone" and "long tone" states and thus learn the correct state representation.

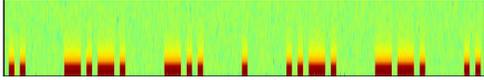

Figure 6: A spectrogram segment of Morse code audio.

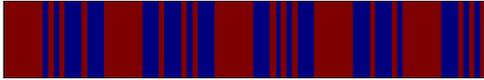

(a) HMM state labeling.

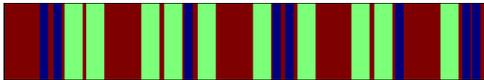

(b) HSMM state labeling.

Figure 7: Each model applied to Morse code data.

In the HDP-HSMM we employ a delayed-geometric duration distribution, in which a state's duration is chosen by first waiting some $w$ samples and then sampling a geometric. Both the wait $w$ and geometric parameter $p$ are learned from data, with a uniform prior over the set $\{0, 1, \ldots, 20\}$ for $w$ and a Beta$(1, 1)$ uniform prior over $p$. This duration distribution class is also a superset of the class of geometric distributions, since the wait parameter $w$ can be learned to be 0.

We applied both the HDP-HSMM and HDP-HMM to the data and found that both quickly concentrate at single explanations: the HDP-HMM finds only two states while the HDP-HSMM correctly disambiguates three, shown in Figure 7. The two "tone" states learned by the HDP-HSMM have $w$ parameters that closely capture the near-deterministic pulse widths, with $p$ learned to be near 1. The "silence" segments are better explained as one state with more variation in its duration statistics. Hence, the HDP-HSMM correctly uncovers the Morse Code alphabet as a natural explanation for the statistics of the audio data.

### 5.3 Speaker Diarization

The NIST Rich Transcription Database is a standard dataset for the speaker diarization problem. It consists of audio recordings for each of 21 meetings with various numbers of participants. The Sticky HDP-HMM of [2] achieved state-of-the-art diarization performance on this dataset using a similar inference scheme. We use this dataset to demonstrate the practical differences in the HDP-HSMM sampling algorithm.

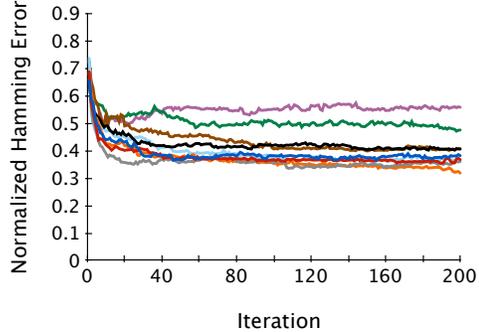

Figure 8: Relatively fast burn-in of an HDP-HSMM sampler. Compare to Figure 3.19(b) of [3].

We used the first 19 Mel Frequency Cepstral Coefficients (MFCCs) computed over 30ms windows spaced every 10ms as our feature vectors, and reduced the dimensionality from 19 to 4 by projecting onto the first four principle components. We also smoothed and subsampled the data so as to make each discrete state correspond to 100ms of real time, resulting in observation sequences of length approximately 8000–10000. We modeled the features as mixtures of Gaussians. Our observation setup mostly follows that of [2], but our time binning is significantly finer. For duration distributions, we again employed the delayed-geometric with the prior on each state's wait parameter as uniform over $\{40, 41, \ldots, 60\}$. In this way we not only impose a minimum duration to avoid rapid state switching or learning in-speaker dynamics, but also force the state sampler to make minimum "block" moves of nontrivial size so as to speed mixing.

Figure 8 shows the progression of nine different HDP-HSMM chains on the NIST_20051102-1323 meeting over a small number of iterations. Within two hundred iterations, most chains have achieved approximately 0.4 normalized Hamming error or less, while it takes 5000 to 30000 iterations for the Sticky HDP-HMM sampler to mix to the same performance on the same meeting, as shown in Figure 3.19(b) of [3]. This reduction in the number of iterations for the sampler to "burn in" more than makes up for the greater computation time per iteration.

We ran 9 chains on each of the 21 meetings to 750 iterations, and Figure 9 summarizes the normalized Hamming distance performance for the final sample of the median chain for each meeting. The performance after the relatively small number of iterations is varied; for some meetings an excellent segmentation with normalized Hamming error around 0.2 is quickly identified, while for other meetings the chains are not able to mix. See Figure 11 for prototypical low-error and high-error meetings, which demonstrates rapid mixing for the meetings on which the model performed well. The low-performance meetings tended to be the same as those with lower performance in [2]. Finally, Figure 10 summarizes the number of inferred speakers compared to the true number of speakers, where we only count speakers whose speech totals at least 5% of the total meeting time.

## 6   Conclusion

The HDP-HSMM is a flexible model for capturing the statistics of non-Markovian data while providing the same Bayesian nonparametric advantages of the HDP-HMM. Furthermore, the sampling algorithms developed here for the HDP-HSMM not only provide relatively fast-mixing inference for the HDP-HSMM, but also produce new algorithms for the original HDP-HMM that warrant further study.

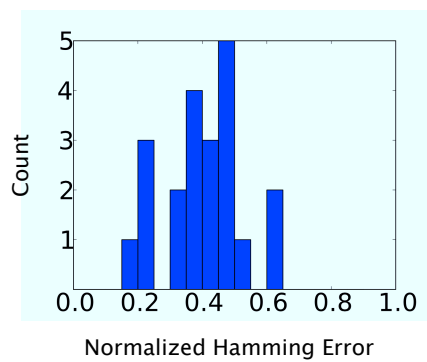

Figure 9: Diarization Performance Summary

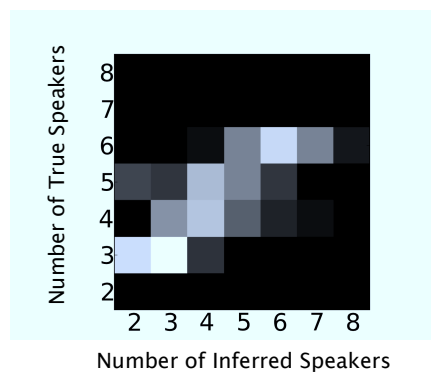

Figure 10: Frequency of Inferred Number of Speakers.

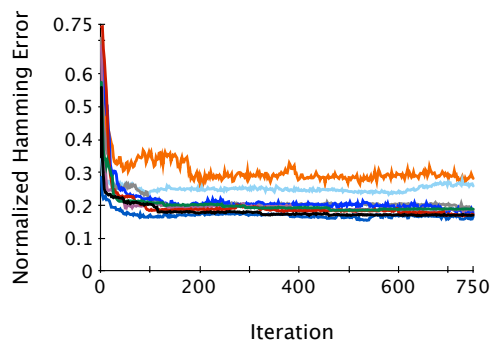

(a) Good-performance meeting.

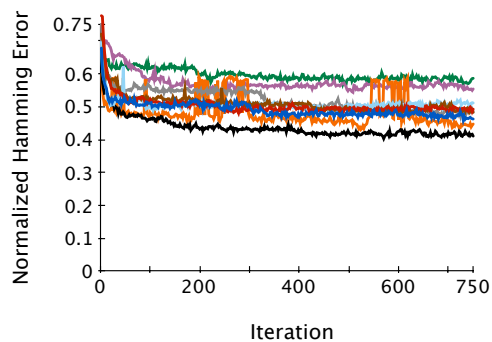

(b) Poor-performance meeting.

Figure 11: Prototypical sampler trajectories for good- and poor-performance meetings.